\documentclass{LTHtwocol} % Use this when you work on your report.
\usepackage{subfiles}
\usepackage{caption}
\usepackage{subcaption}
\usepackage{multirow}
\usepackage{float}
\usepackage{graphicx}
\graphicspath{ {./resources/} }
\usepackage{adjustbox}
\usepackage{pgfgantt}
\usepackage{appendix}
\usepackage[subpreambles=true]{standalone}
\usepackage{import}
\usepackage[utf8]{inputenc}
\usepackage{graphicx,xcolor}
\usepackage{hyperref} 
\hypersetup{
  colorlinks,
  allcolors={blue!40!black},
}
\usepackage{kantlipsum} % Only for the dummy text. Remove for your own report.

\begin{document}
\begin{frontmatter}
\title{Reinforcement Learning-Based Control of CrazyFlie 2.X Quadrotor} % Title of the project.
                      % Note that all reports are in English,
                      %so that our international students can read them.

\author[aj]{Arshad Javeed}
\author[val]{Valentín López Jiménez}

\email[aj]{ar1886ja-s@student.lu.se}
\email[val]{va7764lo-s@student.lu.se}

\advisor[johan]{Johan Grönqvist}
\email[johan]{johan.gronqvist@control.lth.se}

\begin{abstract}
The objective of the project is to explore synergies between classical control algorithms such as PID and contemporary reinforcement learning algorithms to come up with a pragmatic control mechanism to control the CrazyFlie 2.X quadrotor. The primary objective would be performing PID tuning using reinforcement learning strategies. The secondary objective is to leverage the learnings from the first task to implement control for navigation by integrating with the lighthouse positioning system. Two approaches are considered for navigation, a discrete navigation problem using Deep Q-Learning with finite predefined motion primitives, and deep reinforcement learning for a continuous navigation approach. Simulations for RL training will be performed on gym-pybullet-drones, an open-source gym-based environment for reinforcement learning, and the RL implementations are provided by stable-baselines3.
\end{abstract}

\end{frontmatter}
\section{Introduction}

\begin{figure}[H]
	\centering
	\caption{CrazyFlie 2.1}
	\includegraphics[scale=0.5]{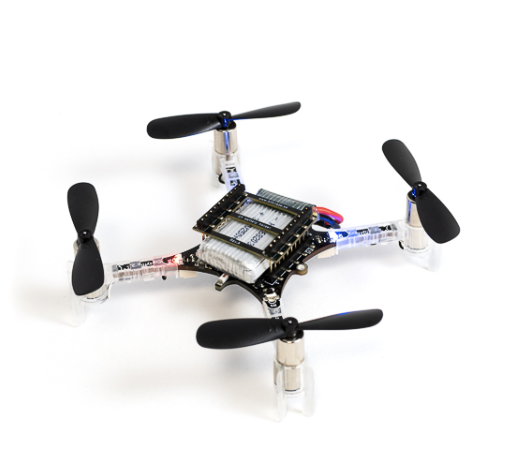}
	\label{cf:fig}
\end{figure}

Modeling a quadrotor such as CrazyFlie (figure \ref{cf:fig}) is not a straightforward task due to the non-linearities involved. Often, the system is linearized around a specific stationary point, but this is task-specific and could be daunting. Instead, we focus on a gray box approach, where we simulate our system using a physics engine \cite{pybullet_gym} to circumvent the physical modeling of the system.
In this paper, we synergize between classical PID control and reinforcement learning is explored to perform a navigation task in the CrazyFlie 2.X quadrotor. Pure classical or reinforcement learning approaches are not feasible in terms of convergence and are less interpretable. A pure RL approach demands a higher network architecture and long training hours to reach convergence. On the other hand, an end-to-end classical approach \cite{modelling_cf}\cite{nonlinear_control} controller's implementations can be complex as the design specification are not trivial.
 
In the research, first, we focus on PID tuning, where the parameters for the attitude and position Mellinger controller \cite{mellinger} are approximated using the Twin-Deep Deterministic Policy Gradient\cite{td3} algorithm and compared against the quadcopter's original values. Next, using the obtained PID parameters, a closed-loop controller is implemented in the simulation environment, where the quadrotor's task is to navigate to a determined point in space in a continuous environment. The RL agent is responsible for the high-level tasks and the PID loop executes the actions. Finally, robustness performance differences are explored in training with and without noise disturbances for the previous hovering navigation task.

 During the implementation of the RL tasks, the algorithm selected is TD3, from \verb|stable-baselines3| \cite{stable_baselines3}, considering its simplicity and robustness. We expect other more advanced actor-critic models to perform similarly.

\section{PID Tuning}

The objective of the task is to determine the PID parameters for the reinforcement learning agent, based on Mellinger's PID architecture which consists of 18 parameters divided into position and attitude controllers (table \ref{pid_tab1:table}. This architecture is already implemented on CrazyFlie's firmware, where the user writes the parameters internally to its memory. 

The following subsections will describe the process of obtaining and validating the parameters from agent-train to hardware test.

\subsection{Agent training}
Agent's PID parameters are trained over a determined number of episodes, starting from random parameters to an optimal solution. The task's main objective is to complete successfully a trajectory, for training a circle and testing a  helix. 

Using TD3 as a reinforcement learning algorithm, convergence has been achieved in  1000 time steps with a small network architecture for the Actor [50,50] and Critic [50,50]. The observation space is defined by position $XYZ$ and orientation $RPY$. The reward function is computed as $(s'-t)^2$ where $t$ is the next position in the target trajectory and $s'$ is the current state. See table \ref{pid_rl_hp:table} for detailed information.

\begin{table}[h]
\caption{Hyperparameters for PID Tuning}
	\label{pid_rl_hp:table}
	\centering
\begin{tabular}{|l|l|}
\hline
\textbf{Hyperparameter}                                                     & \textbf{Value}                                                    \\ \hline \hline
\begin{tabular}[c]{@{}l@{}}Actor Net Arch\\ Critic Net Arch\end{tabular}    & \begin{tabular}[c]{@{}l@{}}{[}50,50{]}\\ {[}50,50{]}\end{tabular} \\ \hline
\begin{tabular}[c]{@{}l@{}}Number of Timesteps\\ Learning Rate\end{tabular} & \begin{tabular}[c]{@{}l@{}}1000\\ 0.001\end{tabular}              \\ \hline
Reward Function                                                             & $-(s'-t)^2$                                        \\ \hline
\begin{tabular}[c]{@{}l@{}}Helix Height\\ Helix Radius\end{tabular}         & \begin{tabular}[c]{@{}l@{}}0.5 m\\ 0.3 m\end{tabular}             \\ \hline
\end{tabular}
\end{table}

\subsection{Hardware implementation}
The same helix trajectory implemented in simulation is used in hardware applying a constant velocity. Crazyflie's flying reference system is achieved using one of two methods, a relative (Flow Deck) and an Absolute (Lighthouse). The absolute position system brings accuracy and stability but depends on external modules in a fixed environment, meanwhile the relative offers mobility without any extra tracking devices, but adds drift and instability to the actions.

\section{Navigation}

The aim of the navigation task is to train a reinforcement-learning policy that learns to navigate the CrazyFlie 2.X quadrotor given the destination coordinates in the specified environment. In contrast to the previous task, where we fed in a sequence/trajectory, here the RL policy is expected to learn to generate the trajectory (actions) on its own. The idea is to formulate a navigation environment and train the model in the simulation bed and then export the model for evaluation on real hardware.

To start things off, we focus on a relatively simple task of hovering the CF2.X quadrotor to get a sense of convergence. Although hovering is something that can be accomplished by classical PID (as done in the previous task), the idea here is to gradually introduce complexities in the environment where a simple PID control would fall short, for instance, introducing a wall and having to maneuver around it or have a dynamic or stochastic environment where the classical path planning algorithms like Dijkstra or A* algorithm can have a hard time accommodating the dynamics. It is also worth noting that the phase space for a system like CF2.X has 12 states, which can further exacerbate the task when employing such algorithms. 

\subsection{Environment}

The objective is simple, the goal is for for the reinforcement-learning agent (CF2.X) is to move to the specified destination specified by a set of coordinates $[x_t, y_t, z_t]$ starting from an initial state $S_0$. The state space/observation space consists of the coordinates, orientation, and linear and angular velocities of the quadrotor. So the state vector comprises of 12 state variables, $S = [x, y, z, r, p, y, v_x, v_y, v_z, w_x, w_y, w_z]$. Given a state $S$, the possible actions constitute moving within a 3D cube $|\Delta x| \le 1, |\Delta y| \le 1, |\Delta z| \le 1$ (a continuous action space). There are several ways of executing the action: i. A pure RL approach, where the policy function outputs the low-level control signals - the 4 motor RPMs. ii. Using an open loop control, where a controller is used to compute the control signals (motor RPMs) to move to execute the action, and the control signals are applied for a fixed number of iterations. iii. Executing a closed loop control, here the RL policy is responsible for predicting the optimal high-level actions $(\Delta x, \Delta y, \Delta z)$ and the actions are successfully executed by the trusted PID controller, relieving the RL policy from having to learn granular controls. Approaches i and ii are supported by \verb|gym-pybullet| out of the box, while approach iii proposed as part of the project is a custom implementation, and was found to outperform in terms of convergence and expected reward.

The RL agent was trained using the Twin-Deep Deterministic Policy Gradient (TD3) implementation from \verb|stable_baselines3|. The actor is responsible of prediction an optimal action ($a$) given a state ($S$), $\pi : S \to a$ and the critic is responsible for predicting the reward ($r$) given a state and an action, $Q : (S, a) \to r$. Table \ref{nav_rl:table} summarizes the variables involved. The reward function is defined as the negative squared error of the current state and the target state (equation \ref{nav_r:eq}). Thus, the objective is to maximize the negative reward (ideally close to 0). Both the actor and critic are deep neural networks, the actor net has a \verb|tanh| output action and the critic has a linear output activation.

\begin{table}[h]
\caption{Navigation Task Variables}
	\label{nav_rl:table}
	\centering
	\begin{adjustbox}{max width=0.45\textwidth}
	\begin{tabular}{|c|c|}
		\hline
		\textbf{Variable} & \textbf{Notation}\\
		\hline \hline
		State Space & $S = [x, y, z, r, p, y, v_x, v_y, v_z, w_x, w_y, w_z]$ \\
		Action Space & $a = [\Delta x, \Delta y, \Delta z] $\\
		Min Action & $a_{min} = [-1, -1, -1] $ \\
		Max Action & $a_{max} = [+1, +1, +1] $ \\
		\hline
	\end{tabular}
	\end{adjustbox}
\end{table}

\begin{equation}
	\label{nav_r:eq}
	r(S, a) = - [(x + \Delta x - x_t)^2 + (y + \Delta y - y_t)^2 + (z + \Delta z - z_t)^2]
\end{equation}

Given a state ($S = [x, y, z]$) and an action ($a = [\Delta x, \Delta y, \Delta z]$), executing the action involves successfully moving to the relative coordinates, i.e. the next state is $S' = (x + \Delta x, y + \Delta y, z + \Delta z)$. To avoid the agent taking long stride while executing the actions, we scale the output of the actor according to equation \ref{nav_a_s:eq}, The scaling factor of 0.05 implies that the agent is restricted to a distance of 0.05 m (relatively) along individual axes. The TD3 algorithm also defines an action noise for exploration and better convergence. We define the action noise as a multivariate Gaussian (equation \ref{nav_a_n:eq}), so the effective action is then $a = 0.05 (a + a_n)$.

\begin{equation}
	\label{nav_a_s:eq}
	a = 0.05 * [\Delta x, \Delta y, \Delta z]
\end{equation}

\begin{equation}
	\label{nav_a_n:eq}
	a_n \sim \frac{\exp(-\frac{1}{2} (a - \mu)^T \Sigma (a - \mu))}{\sqrt{(2 \pi)^3 |\Sigma|}}
\end{equation}

Table \ref{nav_rl_hp:table} lists the hyperparameters for training. The control frequency implies the number of control actions during a period of 2 simulation secs to execute the corresponding action.

\begin{table}[h]
\caption{Hyperparameters}
	\label{nav_rl_hp:table}
	\centering
	\begin{tabular}{|c|c|}
		\hline
		\textbf{Hyperparameter} & \textbf{Value} \\
		\hline \hline
		Actor Net Arch & $[50, 100, 500, 100, 50, 3]$ \\
		Critic Net Arch & $[50, 100, 500, 100, 50, 1]$ \\
		\hline
		Number of Timesteps & $100\,000$ \\
		Learning Rate & $0.001$ \\
		\hline
		Action Noise Mean ($\mu$) & $[0, 0, 0]$ \\
		Action Noise Variance ($\Sigma$) & $ \begin{bmatrix}
								   0.5 & 0 & 0 \\ 
								   0 & 0.5 & 0 \\ 
								   0 & 0 & 0.5
								 \end{bmatrix} $ \\
		\hline
		Control Frequency & 50 Hz \\
		\hline
		Initial State (of the drone) & $S_0 = [0, 0, 0]$ \\
		\hline
	\end{tabular}
\end{table}

\section{Robustness}

Assessing the robustness of black box models has always been a challenge. While the control algorithms employing the classical control principles have an empirical to quantify robustness and evaluating performance, quantifying the robustness in the case of black box models relies on deliberate disturbance and adversarial techniques. We resort to the approach of injecting disturbance to evaluate the RL policy and also explore the possibility of improving robustness by subjecting the agent to external disturbance during the training phase. At first glance, it might be reasonable to expect that training with disturbance would make the RL policy more resilient to external disturbances. However, the experimental results refute the assumption. We find the RL agent to have inherent robustness and training with external disturbance did not have a significant impact. Our findings corroborate similar results reported for other continuous control systems \cite{charac_robustness}.

\begin{figure}[H]
	\centering
	\caption{External Disturbances Applied - Training}
	\includegraphics[scale=0.2]{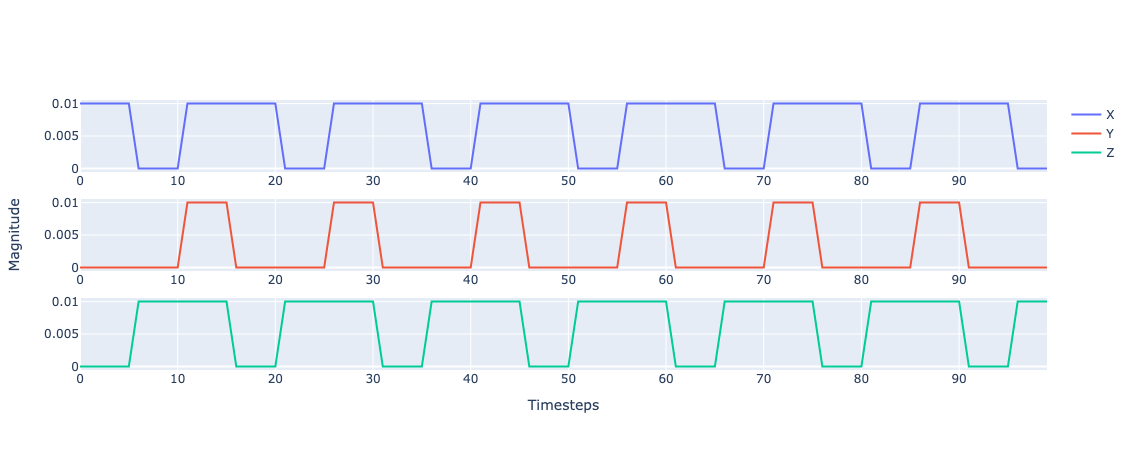}
	\label{rob_ext_dists:fig}
\end{figure}

We focus on step/pulse disturbance, as it is more realistic and emulates the wind disturbance experienced by the quadcopter. However, the disturbance is applied in multiple directions (along the XYZ axes). To make it challenging, during the training phase, the direction of the disturbances is switched every few iterations randomly. Figure \ref{rob_ext_dists:fig} shows the training disturbances applied, the disturbance is applied along X, then Z, and then along all three axes XYZ. The evaluation is based on the fixed step disturbances and the performance is measured individually for varying magnitudes of disturbance along individual axes. This also ensures that the replay buffer always contains a good sample.

\subsection{Hardware Implementation}

Once the reinforcement learning was done, the actor-network (neural network) was extracted. The real-time inference was achieved by running the model on the computer and sending the actions to CrazyFlie 2.1 over the radio by continuously reading the sensor measurements. The compatibility of states was ensured by reading all of the required measurements from the onboard sensors and converting the quantities to respective units. The position ($[x, y, z]$), orientation ($[r, p, y]$) and linear velocities ($[v_x, v_y, v_z]$) were read from the Kalman filter estimates. And the angular velocities ($[\omega_x, \omega_y, \omega_z]$) were obtained from the gyro.

\section{Results}
\subsection{PID}

\begin{figure}[H]
	\centering
	\caption{Helix Test Results in Gym-Pybullet-Drones}
	\begin{subfigure}[b]{0.2\textwidth}
		\caption{X}
		\includegraphics[width=\textwidth]{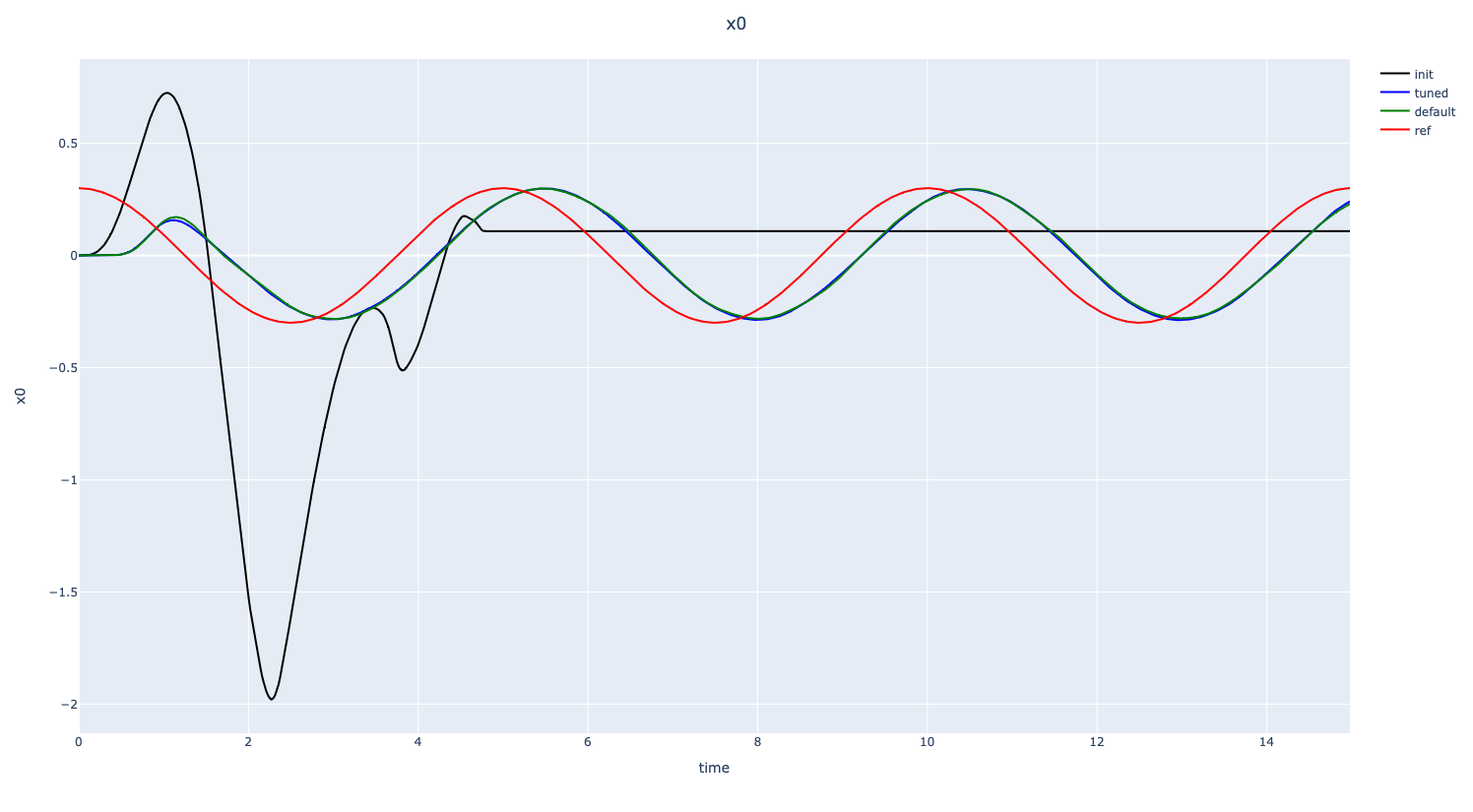}
		\label{fig:figpid1}
	\end{subfigure}
	\hfill
	\begin{subfigure}[b]{0.2\textwidth}
		\caption{Y}
		\includegraphics[width=\textwidth]{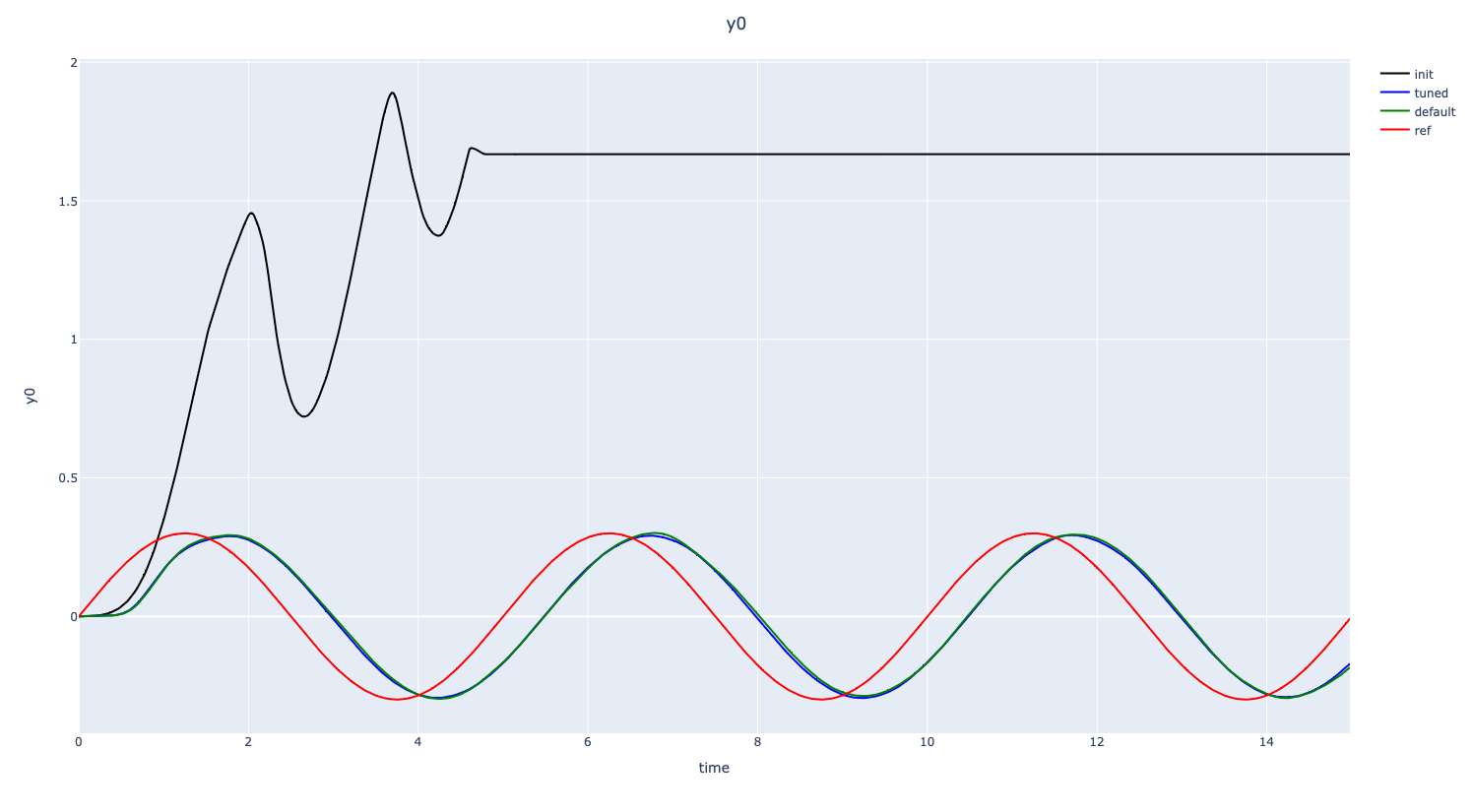}
		\label{fig:figpid2}
	\end{subfigure}
	\hfill
	\begin{subfigure}[b]{0.2\textwidth}
		\caption{Z}
		\includegraphics[width=\textwidth]{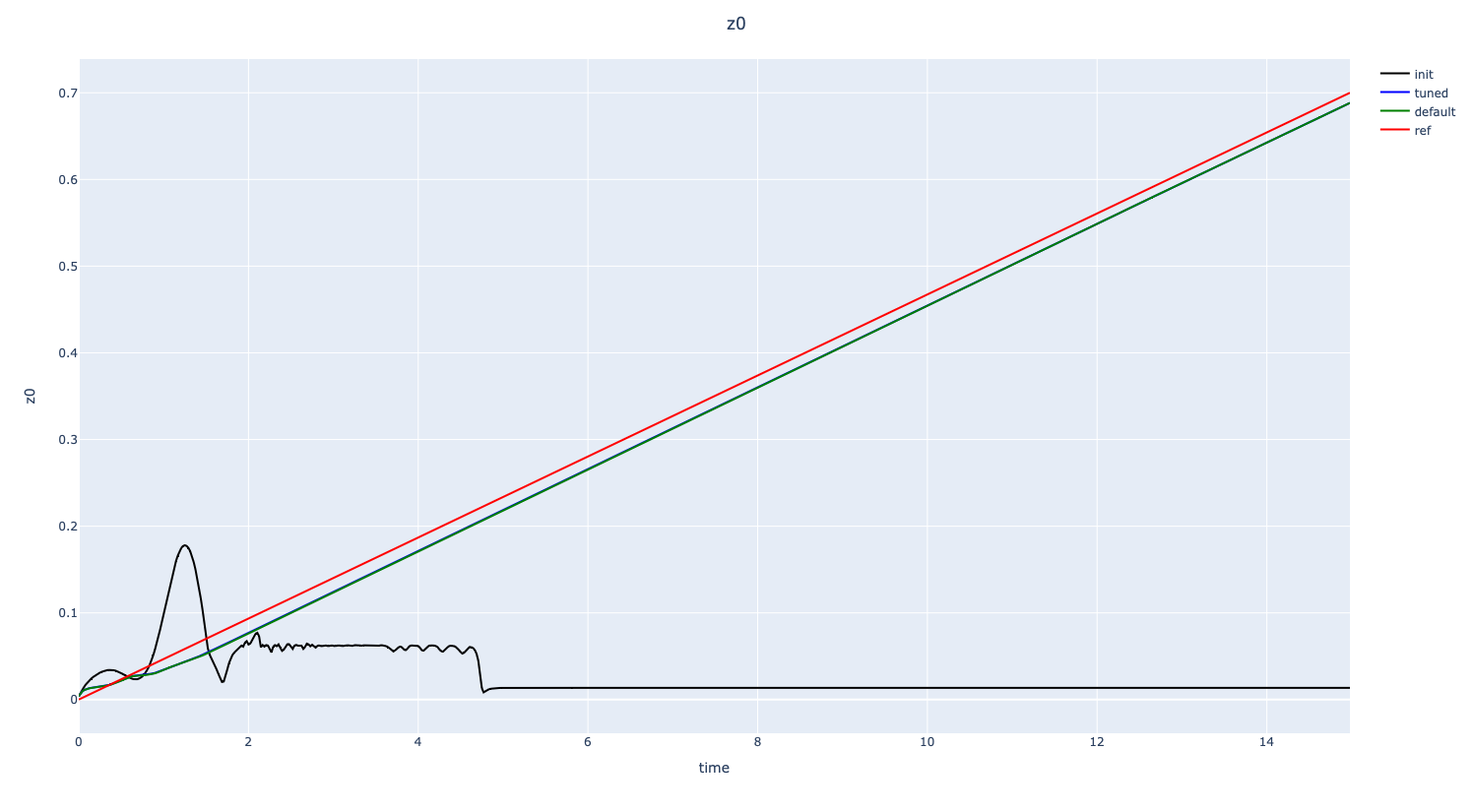}
		\label{fig:figpid3}
	\end{subfigure}
	\hfill
	\begin{subfigure}[b]{0.2\textwidth}
		\caption{Step Response}
		\includegraphics[width=\textwidth]{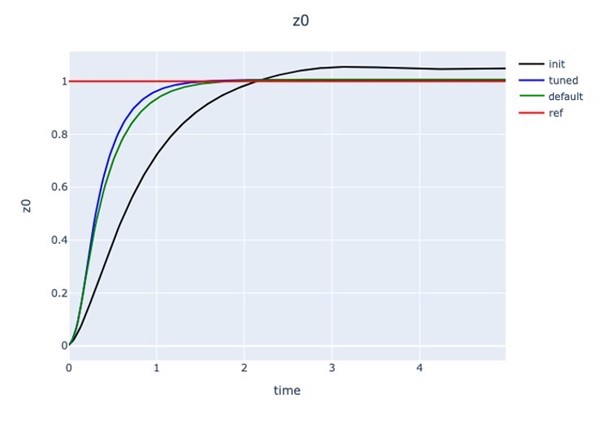}
		\label{pid_step_response:fig}
	\end{subfigure}
\end{figure}

The objective of the agent is to complete a full 360 degrees helix. In the simulation environment, a circle is used for training, and convergence is measured against the Crazyflie trajectory to trace a helix. In figures \ref{fig:figpid1}, \ref{fig:figpid2}, and \ref{fig:figpid3} the output plot compares the tuned (output parameters from the algorithm) vs the default for test trajectory along X, Y and Z axis.

\begin{figure*}[h]
	\centering
	\caption{Helix - Hardware Results}
	\begin{subfigure}[b]{0.4\textwidth}
		\caption{Flow Deck (Relative Positioning)}
		\includegraphics[width=\textwidth]{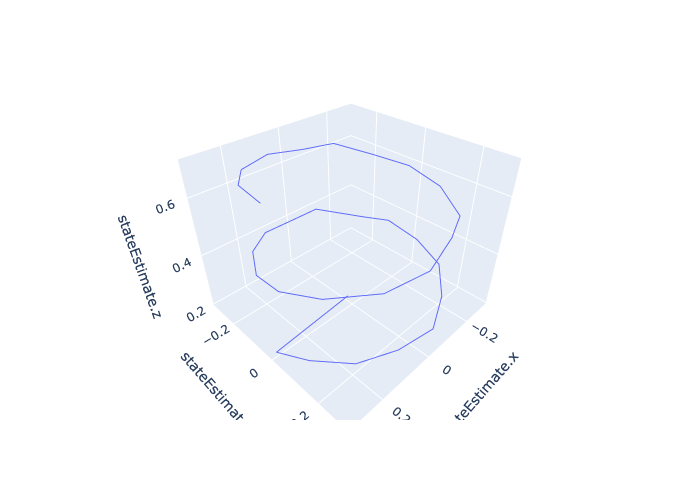}
		\label{fig:figpid4}
	\end{subfigure}
	\hfill
	\begin{subfigure}[b]{0.4\textwidth}
		\caption{Lighthouse Deck (Absolute Positioning)}
		\includegraphics[width=\textwidth]{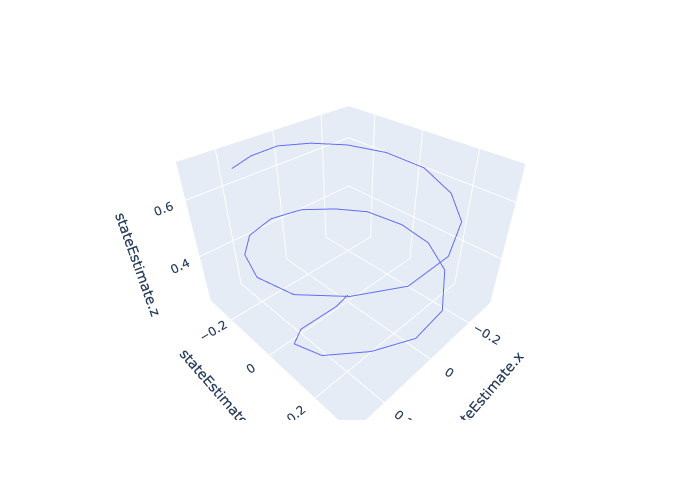}
		\label{fig:figpid5}
	\end{subfigure}
\end{figure*}

Hardware testing includes the two methods mentioned above. First, the relative positioning system uses the Flow deck, which shows a rougher point-to-point displacement, meanwhile, the absolute positioning using the Lighthouse shows a smoother performance. See figures \ref{fig:figpid4} and \ref{fig:figpid5} for more details.

Finally, figure \ref{pid_step_response:fig} shows the step responses, where tuned and default parameters have similar results. Therefore, we can conclude that the estimated model is approximated enough to the real, for more details table \ref{pid_step:table} presents more step response information.

\begin{table*}[h]
\caption{Step Response Characteristics - Randomly initialized vs Tuned vs Default }
\label{pid_step:table}
\centering
\begin{tabular}{|l|c|c|c|}
\hline
\textbf{Characteristics} & \begin{tabular}[c]{@{}l@{}} \textbf{Randomly Initialized} \\ \textbf{Params}\end{tabular} & \textbf{Tuned Params} & \textbf{Default Params} \\ \hline \hline
Rise Time (s)      & 1.557                                                                  & 0.649        & 0.7728         \\ \hline
Settling Time (s)   & 2.445                                                                  & 1.199        & 1.4028         \\ \hline
Overshoot       & 0.548                                                                  & 0.0373       & 0.1901         \\ \hline
Peak            & 1.0546                                                                 & 1.000        & 1.0067         \\ \hline
\end{tabular}
\end{table*}

\begin{table*}[h]
\caption{PID Tuning - Default vs Tuned Coefficients}
\label{pid_tab1:table}
\centering
\begin{tabular}{|c|l|c|c|c|c|c|c|}
\hline
\multirow{3}{*}{\textbf{Position Controller}} &                  & \textbf{kp\_xy} & \textbf{kp\_z} & \textbf{ki\_xy}    & \textbf{ki\_z}    & \textbf{kd\_xy} & \textbf{kd\_z} \\ \cline{2-8} 
                                              & \textbf{Default} & 0.4             & 1.25           & 0.05               & 0.05              & 0.2             & 0.5            \\ \cline{2-8} 
                                              & \textbf{Tuned}   & 0.364           & 1.169          & 0.052              & 0.052             & 0.234           & 0.586          \\ \hline
\multirow{3}{*}{\textbf{Attitude Controller}} & \textbf{}        & \textbf{kR\_xy} & \textbf{kR\_z} & \textbf{ki\_m\_xy} & \textbf{ki\_m\_z} & \textbf{kw\_xy} & \textbf{kw\_z} \\ \cline{2-8} 
                                              & \textbf{Default} & 70000           & 60000          & 0.0                & 500               & 20000           & 12000          \\ \cline{2-8} 
                                              & \textbf{Tuned}   & 64786.842       & 55531.579      & 0.0                & 599.666           & 17217.406       & 10330.444      \\ \hline
\end{tabular}
\end{table*}

\subsection{Navigation}

The objective for the RL agent is to hover at $S_t = [0, 0, 1]$ starting from the origin $S_0 = [0, 0, 0]$. Throughout the training iterations, the starting point remains fixed. Once trained, we evaluate the RL policy in a similar setting to the training task, i.e. starting at the origin, but also test the ability to generalize by changing the starting point to an arbitrary location which was never explored by the agent during training, we also subject agent to external disturbances (wind) while evaluating the model in real-world.

\begin{figure*}[h]
	\centering
	\caption{RL Training Performance: Comparison of RL + Open Loop Control and RL + Closed Loop Control}
	{Figures plotting the RL results for the task of hovering using open loop control to execute the actions. The X-axis represents the relative timesteps. The Y-axis represents the magnitude. Total number of timesteps for RL + open loop system is 1\,000\,000 (1 M), and 100\,000 (100 K) for the RL + closed loop system. The performance metric used here is the expected reward per step (higher the better)}
	\includegraphics[scale=0.4]{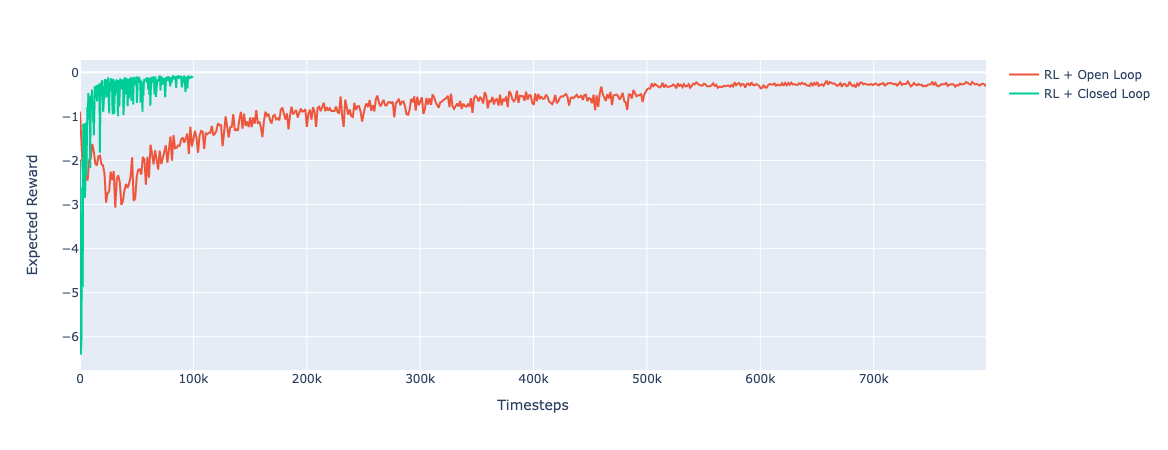}
	\label{nav_hov_oc:fig}
\end{figure*}

The results are presented and compared for the three approaches described before i. Pure RL. ii. RL with open loop control. iii. RL with closed loop control. Table \ref{rl_ts:table} summarizes the results. The pure RL approach (directly controlling the motor RPMs) was not found to converge even after training for 10 million steps. The RL with the open loop control approach managed to converge after ~10 hrs of training, while the custom implementation of RL and a PID loop managed to converge in significantly less training time and achieves a better-expected reward (reward per step in the episode). Figure \ref{nav_hov_oc:fig} compare the results of approaches (ii) and (iii). We see that the agent in approach (iii) learns significantly faster and manages to attain a higher reward early in the training phase and also attains a higher expected reward at the end of training. This is due to the fact that the agent has robust low-level control to execute the necessary motion primitives (using a closed loop PID controller) and does not have to learn them from scratch, unlike approach (ii). It is also worth noting that the magnitude of expected reward in (iii) is quite high compared to (ii), a negative reward of -6 vs -3 (per step), this is again due to the fact that we leverage robust low-level control, because of which the agent can execute actions concretely and explore more of the environment.
The actor reward after convergence is close to about 1, which is expected, as the shortest distance between $S_0 = [0, 0, 0], S_t = [0, 0, 1]$ is 1 (Euclidean distance).

\begin{figure}[H]
	\centering
	\caption{RL Policy Test}
	{The plots trace the movements of the agent in the environment. Red indicates the starting point and green is the destination.}

	\begin{subfigure}[b]{0.25\textwidth}
		\caption{Case 1: Starting at origin (same as training task)}
		\includegraphics[width=\textwidth]{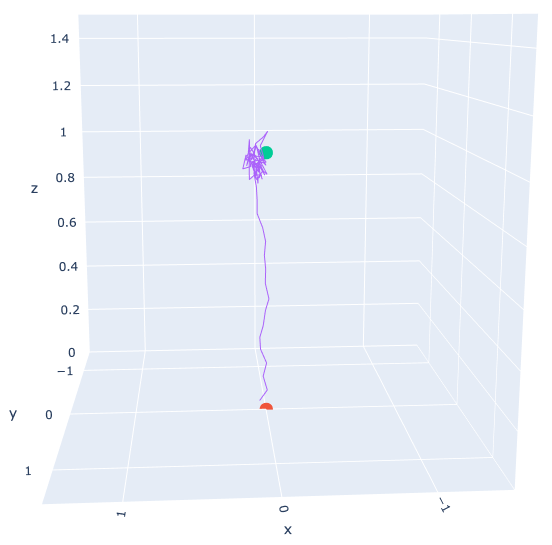}
		\label{rl_c_3d-plot000:fig}
	\end{subfigure}
	\hfill
	\begin{subfigure}[b]{0.25\textwidth}
		\caption{Case 2: Starting at an arbitrary location (trajectory not explored during training)}
		\includegraphics[width=\textwidth]{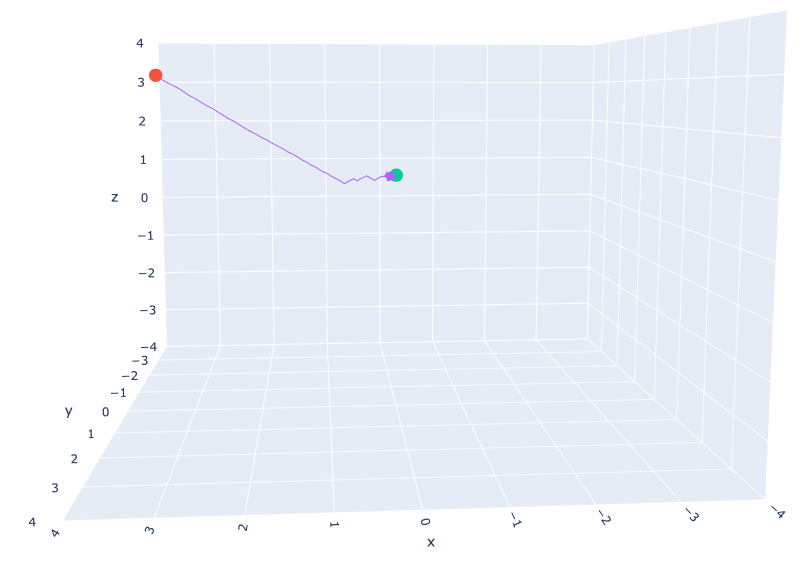}
		\label{rl_c_3d-plot333:fig}
	\end{subfigure}
\end{figure}

Approach (iii) was simulated after training converged. Figure \ref{rl_c_3d-plot000:fig} is the same as the training task, with the starting point being the origin. In figure \ref{rl_c_3d-plot333:fig}, the starting point was chosen arbitrarily, $S_0=[3, 3, 3]$, which was never explored by the agent during the training phase, yet the policy generalizes well, but the subtle thing to note is that the trajectory is not entirely optimal (the shortest path between two points is a straight line), instead the agent first navigates to a location close to the destination and then pursues a familiar trajectory encountered during training.

The model was also tested to control the real hardware and the results were similar. While testing on the real hardware, external disturbances were applied in attempts to drift the CF2.X from the hovering point, and the model could stabilize fairly stabilize and return to the hovering point.

\begin{table}[H]
\caption{Training Summary}
\label{rl_ts:table}
\centering
\begin{adjustbox}{max width=0.45\textwidth}
\begin{tabular}{|l|c|c|c|}
\hline
\multicolumn{1}{|c|}{\textbf{Metrics}} & \multicolumn{3}{c|}{\textbf{Approach}}                                                                               \\ \hline
\multicolumn{1}{|c|}{\textbf{}}        & \multicolumn{1}{c|}{\textbf{Pure RL}}         & \multicolumn{1}{c|}{\textbf{RL + Open loop}} & \textbf{RL + Closed loop} \\ \hline
\multicolumn{1}{|l|}{\textbf{Description}}          & \multicolumn{1}{c|}{\begin{tabular}[c]{@{}c@{}}No controller needed, the RL agent\\  learns to execute the \\ low-level actions on its own.\end{tabular}} & \multicolumn{1}{c|}{\begin{tabular}[c]{@{}c@{}}The open-loop makes the RL\\  task easier, but the execution \\ of actions is not consistent.\end{tabular}} & \multicolumn{1}{c|}{\begin{tabular}[c]{@{}c@{}}Abstracts all of the low-level control\\  from RL and ensures consistency\\  in action execution\end{tabular}} \\ \hline
Number of Timesteps                    & \multicolumn{1}{c|}{\textgreater 10 Million} & \multicolumn{1}{c|}{$\sim$1 Million}         & $\sim$100 K            \\ \hline
Training Time $\dag$                   & \multicolumn{1}{c|}{\textgreater 13 hrs}     & \multicolumn{1}{c|}{$\sim$10 hrs}            & $\sim$1.5 hrs          \\ \hline
Expected Reward                        & \multicolumn{1}{c|}{-}                       & \multicolumn{1}{c|}{-0.4}                    & -0.1                   \\ \hline
Convergence                            & \multicolumn{1}{c|}{No}                      & \multicolumn{1}{c|}{Yes}                     & Yes                    \\ \hline
\multicolumn{4}{l}{\footnotesize $\dag$ all training durations are reported on the same hardware}\\
\end{tabular}
\end{adjustbox}
\end{table}

\subsection{Roubstness}

\begin{figure}[H]
	\centering
	\caption{Robustness Evaluation - External Disturbance Injection}
	{Comparing expected rewards of RL policy trained without and with external steps disturbance}
	\begin{subfigure}[b]{0.45\textwidth}
		\caption{Distrubance Along X}
		\includegraphics[width=\textwidth]{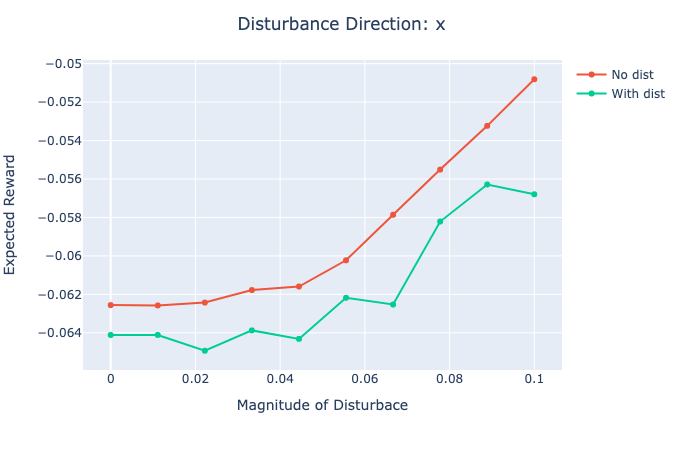}
		\label{dist_x:fig}
	\end{subfigure}
	\hfill
	\begin{subfigure}[b]{0.45\textwidth}
		\caption{Distrubance Along Z}
		\includegraphics[width=\textwidth]{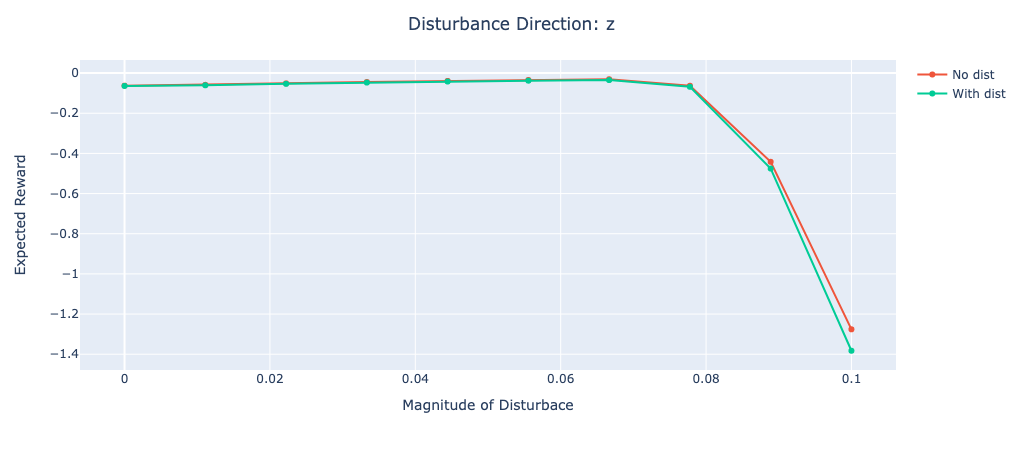}
		\label{dist_z:fig}
	\end{subfigure}
	\hfill
	\begin{subfigure}[b]{0.45\textwidth}
		\caption{Distrubance Along XYZ}
		\includegraphics[width=\textwidth]{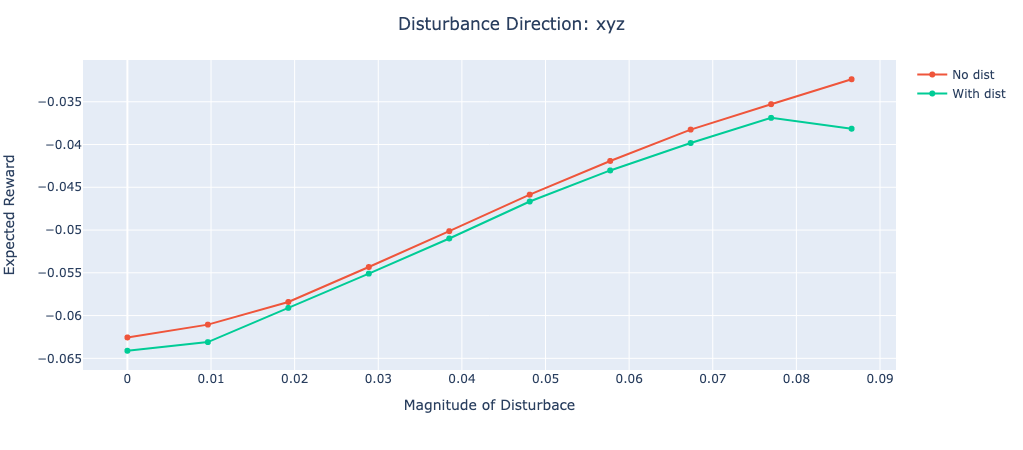}
		\label{dist_xyz:fig}
	\end{subfigure}
\end{figure}

Figures \ref{dist_x:fig} - \ref{dist_xyz:fig} compare the performance of the model trained with and without the disturbance. During the evaluation run, the disturbance vector is held constant throughout the episode (step disturbance). The magnitude of the disturbance is varied along with the directions and the expected rewards are recorded individually.

When a disturbance is applied along X or XYZ (figures \ref{dist_x:fig} \& \ref{dist_xyz:fig}), the expected reward appears to increase while applying the disturbance along Z (figure \ref{dist_z:fig}) seems to have a negative impact. A disturbance along XY direction appears to tilt the quadrotor and it appears to leverage the combined effect of the actions taken and the XY component of the wind force, making it approach the hovering point much faster, thus a better reward. This explains why a disturbance along Z has a drastic negative impact, the wind is constantly pushing the quadrotor upwards, and when the magnitude gets large, the quadrotor has to work against the wind. We also observe that the model trained without any disturbance attains a better reward when evaluating disturbances along XY direction. In the case of disturbance along Z, both models seem to perform similarly, and as expected the expected reward decreases with the magnitude of noise, because the agent cannot use the wind force along Z to its advantage, instead it was to work against it.

The robustness evaluation reveals the model trained without any disturbance during the training phase is inherently robust to external disturbances when subjected to wind forces during the testing phase. Not only the model exhibits some form of robustness, but in fact outperforms the model that was subjected to random step disturbances during training. It was also observed that during training injecting a higher magnitude of disturbance resulted in the model not converging, going unstable as expected.
 A plausible explanation is that a single RL agent model is unable to learn/differentiate between the behavior of the combined system, i.e. the drone plus the environment dynamics. Our observations corroborate similar findings reported by a similar study for the cart pole experiment\cite{charac_robustness}. Thus we conclude that training a model subject to external disturbance does not have a significant impact on the model's robustness, but it does help in exploring the environment faster.

\begin{figure}[H]
	\centering
	\caption{3D trajectories against disturbances for final model}
	{Robustness against disturbances in simulation and hardware}
	\begin{subfigure}[b]{0.25\textwidth}
		\caption{Distrubance Along X in simulation}
		\includegraphics[width=\textwidth]{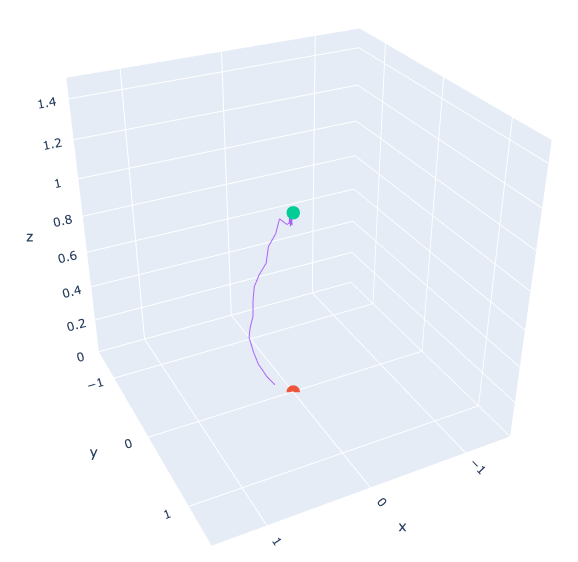}
		\label{final_x:fig}
	\end{subfigure}
	\hfill
	\begin{subfigure}[b]{0.25\textwidth}
		\caption{Distrubance Along XYZ in simulation}
		\includegraphics[width=\textwidth]{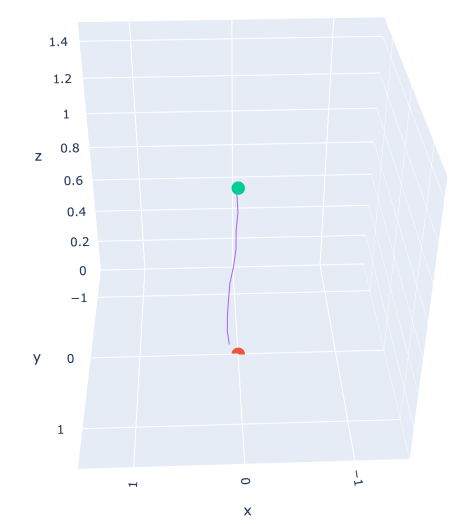}
		\label{final_xyz:fig}
	\end{subfigure}
	\hfill
	\begin{subfigure}[b]{0.25\textwidth}
		\caption{Distrubances along XYZ on Hardware}
		\includegraphics[width=\textwidth]{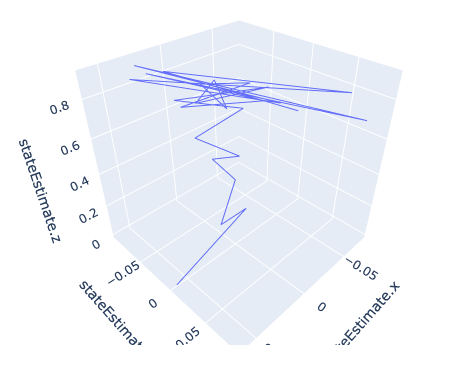}
		\label{final_hardware:fig}
	\end{subfigure}
\end{figure}

\begin{figure}[H]
	\centering
	\caption{Final model hardware test}
	\includegraphics[scale=0.3]{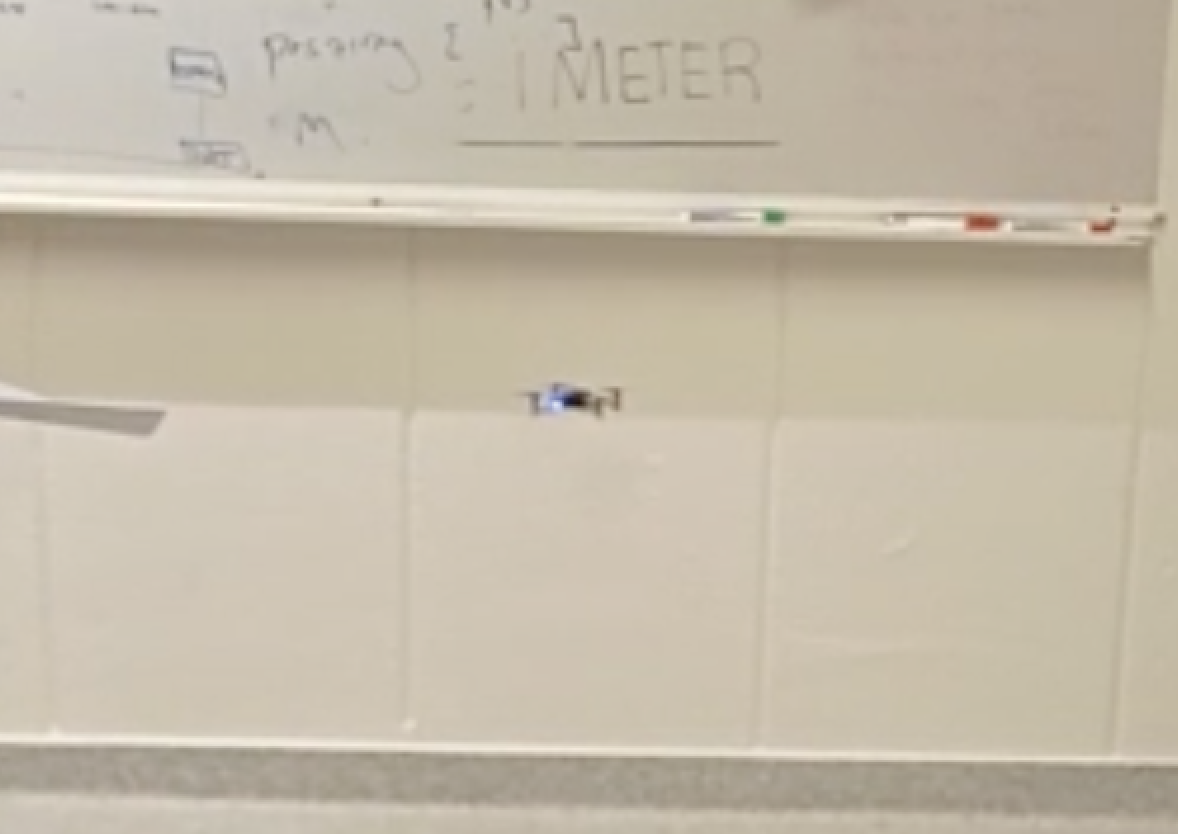}
	\label{hardware:fig}
\end{figure}

Figures \ref{final_x:fig} - \ref{final_xyz:fig} are the simulated 3D trajectories, that show the model stable, reaching the target point $[0,0,1]$. Behavior that is confirmed in hardware with figures \ref{final_hardware:fig} and \ref{hardware:fig}, where the CrazyFlie is subject to disturbances in different directions (wind force exerted manually).

\section{Conclusion}

The project focused on reinforcement learning-based control of a CrazyFlie, exploring aspects of combining classical control and reinforcement learning approaches. Our first objective was to determine how suitable was this virtual environment, choosing as a primary task, the PID tuning of the controller coefficients. In order to do so, the agent needed to complete a predefined trajectory in simulation and reality. This formed the basis for the navigation task, where the RL agent was responsible for predicting high-level actions for maneuvers, and the PID controller was leveraged to execute the actions successfully. 
	
	The navigation task also explores 3 distinct approaches, i. A pure RL approach, where RL agent directly outputs the low-level control signal (without the PID controller). ii. RL and open-loop approach, where the agent predicts the actions and the low-level PID control computes the control signal and executes them in an open-loop fashion. iii. Finally, the RL and closed-loop approach was improved to run a closed-loop PID to ensure the high-level actions were perfectly executed before moving on to the next RL step. We observe that approach (iii) performed remarkably, achieving the best results. Approach (iii) aims to borrow the best from both fields, robust and explainable principles (PID) from classical control, and the ability to adapt and perform complex tasks from reinforcement learning.
	
	Finally, as part of robustness evaluation, we turn towards assessing the inherent robustness of RL models and explore if subjecting the agent to external disturbances would improve the performance. The main objective was to establish a comparison of how an agent trained against disturbances could improve the model performance. In our case, the model subject to disturbance showed poor performance. We conclude that for our experiment the RL agents have inherent robustness and training with disturbances does not improve the performance, but it does help the agent explore the environment.

\section{Acknowledgements}

We acknowledge our project advisors Johan Grönqvist and Emma Tegling for all the guidance and support. We would also like to acknowledge the Department of Automatic Control, Lund University for all the resources.

\newpage
\printbibliography

@article{stable_baselines3,
  author  = {Antonin Raffin and Ashley Hill and Adam Gleave and Anssi Kanervisto and Maximilian Ernestus and Noah Dormann},
  title   = {Stable-Baselines3: Reliable Reinforcement Learning Implementations},
  journal = {Journal of Machine Learning Research},
  year    = {2021},
  volume  = {22},
  number  = {268},
  pages   = {1-8},
  url     = {http://jmlr.org/papers/v22/20-1364.html}
}

@inproceedings{mellinger,
  author={Mellinger, Daniel and Kumar, Vijay},
  booktitle={2011 IEEE International Conference on Robotics and Automation}, 
  title={Minimum snap trajectory generation and control for quadrotors}, 
  year={2011},
  volume={},
  number={},
  pages={2520-2525},
  doi={10.1109/ICRA.2011.5980409}}

@inproceedings{pybullet_gym,
      title={Learning to Fly---a Gym Environment with PyBullet Physics for Reinforcement Learning of Multi-agent Quadcopter Control}, 
      author={Jacopo Panerati and Hehui Zheng and SiQi Zhou and James Xu and Amanda Prorok and Angela P. Schoellig},
      booktitle={2021 IEEE/RSJ International Conference on Intelligent Robots and Systems (IROS)},
      year={2021},
      volume={},
      number={},
      pages={},
      doi={}
}

@inproceedings{
charac_robustness,
title={Characterising the Robustness of Reinforcement Learning for Continuous Control using Disturbance Injection},
author={Catherine Glossop and Jacopo Panerati and Amrit Krishnan and Zhaocong Yuan and Angela P. Schoellig},
booktitle={Progress and Challenges in Building Trustworthy Embodied AI},
year={2022},
url={https://openreview.net/forum?id=IgXOXUVObLB}
}

@article{td3,
  author       = {Scott Fujimoto and
                  Herke van Hoof and
                  David Meger},
  title        = {Addressing Function Approximation Error in Actor-Critic Methods},
  journal      = {CoRR},
  volume       = {abs/1802.09477},
  year         = {2018},
  url          = {http://arxiv.org/abs/1802.09477},
  eprinttype    = {arXiv},
  eprint       = {1802.09477},
  bibsource    = {dblp computer science bibliography, https://dblp.org}
}

@misc{modelling_cf,
  author       = {{Greiff, Marcus}},
  issn         = {{0280-5316}},
  language     = {{eng}},
  note         = {{Student Paper}},
  title        = {{Modelling and Control of the Crazyflie Quadrotor for Aggressive and Autonomous Flight by Optical Flow Driven State Estimation}},
  year         = {{2017}},
}

@phdthesis{nonlinear_control,
  author       = {{Greiff, Marcus}},
  isbn         = {{978-91-8039-047-7}},
  language     = {{eng}},
  month        = {{10}},
  publisher    = {{Department of Automatic Control, Lund University}},
  school       = {{Lund University}},
  title        = {{Nonlinear Control of Unmanned Aerial Vehicles : Systems With an Attitude}},
  url          = {{https://lup.lub.lu.se/search/files/109517053/MG_thesis_final.pdf}},
  year         = {{2021}},
}

\onecolumn
\newpage

%\onecolumn
\section*{Appendix A: Software/Tools Used}

\begin{table}[H]
\label{app_a_software:table}
\centering
\begin{tabular}{|l|l|l|l|}
\hline
\textbf{Library}    & \textbf{Description}                                                                                                                        & \textbf{Version} & \textbf{License} \\ \hline
gym                 & A universal API for reinforcement learning environments                                                                                     & 0.21.0           & MIT License      \\ \hline
gym-pybullet-drones & \begin{tabular}[c]{@{}l@{}}PyBullet Gym environments for single and multi-agent\\ reinforcement learning of quadcopter control\end{tabular} & 0.0.3            & MIT License      \\ \hline
PyTorch             & \begin{tabular}[c]{@{}l@{}}Tensors and Dynamic neural networks in Python\\ with strong GPU acceleration\end{tabular}                        & 1.11.0           & BSD-3            \\ \hline
stable-baselines3   & \begin{tabular}[c]{@{}l@{}}Pytorch version of Stable Baselines, implementations of \\ reinforcement learning algorithms\end{tabular}        & 1.8.0            & MIT License      \\ \hline
cflib               & Crazyflie python driver                                                                                                                     & 0.1.22           & GPLv3            \\ \hline
\end{tabular}
\end{table}

%\onecolumn
\section*{Appendix B: Hardware}

\begin{table}[H]
\label{app_b_hardware:table}
\centering
\begin{tabular}{|l|l|l|}
\hline
\textbf{Hardware}          & \textbf{Description}                                                                                      & \textbf{Version} \\ \hline
CrazyFlie                  & Mini quadrotor                                                                                            & 2.1              \\ \hline
Caryradio                  & USB dongle for comm over radio                                                                            & 2.0              \\ \hline
Flow deck                  & Relative positioning system                                                                               & 2.0              \\ \hline
Lighthouse Deck            & Absolute positioning system                                                                               &                  \\ \hline
2 Lighthouse Base stations & \begin{tabular}[c]{@{}l@{}}Enables on-board positioning together\\  with the lighthouse deck\end{tabular} & 2.0              \\ \hline
\end{tabular}
\end{table}

%\section{Plan}
%
%\begin{ganttchart}[
%vgrid={draw=none, dotted},
%bar/.append style={fill=black},
%expand chart=\textwidth
%]{1}{11}
%    \gantttitle{Week}{11} \\
%    \gantttitlelist{1,...,11}{1} \\
%    \ganttbar[progress=100]{4.1.a: PID Tuning (Sim)}{1}{2} \\
%    \ganttbar[progress=100]{4.1.b: Demo 1: PID}{3}{3} \\
%    \ganttbar[progress=100]{4.2.d: Lighthouse Positioning}{3}{4} \\
%    \ganttbar[progress=100]{4.2.a: Env for Nav (Sim)}{4}{4} \\
%    \ganttbar{4.2.b: Discrete Problem (Sim)}{5}{6} \\
%    \ganttbar[progress=50]{4.2.c: Continuous Problem (Sim)}{5}{6} \\
%    \ganttbar[progress=0]{4.2.e: *** Explore AI deck}{10}{11} \\
%    \ganttbar[progress=50]{4.2.f: Demo 2: Comm over Radio}{7}{8} \\
%    \ganttbar[progress=25]{4.2.g: ** Robustness/Obstacle Avoidance}{8}{9} \\
%\end{ganttchart}

\end{document}